# THE DESIGN AND IMPLEMENTATION OF LANGUAGE LEARNING CHATBOT WITH XAI USING ONTOLOGY AND TRANSFER LEARNING

Nuobei SHI, Qin Zeng and Raymond Lee

Division of Science and Technology, Beijing Normal University-Hong Kong Baptist University United International College, Zhuhai, China

### ABSTRACT

*In this paper, we proposed a transfer learning-based English language learning chatbot, whose output generated by GPT-2 can be explained by corresponding ontology graph rooted by fine-tuning dataset. We design three levels for systematically English learning, including phonetics level for speech recognition and pronunciation correction, semantic level for specific domain conversation, and the simulation of "free-style conversation" in English - the highest level of language chatbot communication as 'free-style conversation agent'. For academic contribution, we implement the ontology graph to explain the performance of free-style conversation, following the concept of XAI (Explainable Artificial Intelligence) to visualize the connections of neural network in bionics, and explain the output sentence from language model. From implementation perspective, our Language Learning agent integrated the mini-program in WeChat as front-end, and fine-tuned GPT-2 model of transfer learning as back-end to interpret the responses by ontology graph.*

*All of our source codes have uploaded to GitHub:*
*https://github.com/p930203110/EnglishLanguageRobot.*

## 1. INTRODUCTION

*Language chatbot* has widely used in customer services or personal assistants for task-orientated, interactive chats in special domains and knowledge base for question-answer systems. All have comprised of automatic speech recognition (ASR), natural language understanding (NLU), dialogue management (DM), natural language generation (NLG), speech synthesis (SS). Figure 1 shows the system flow of a typical chatbot system.

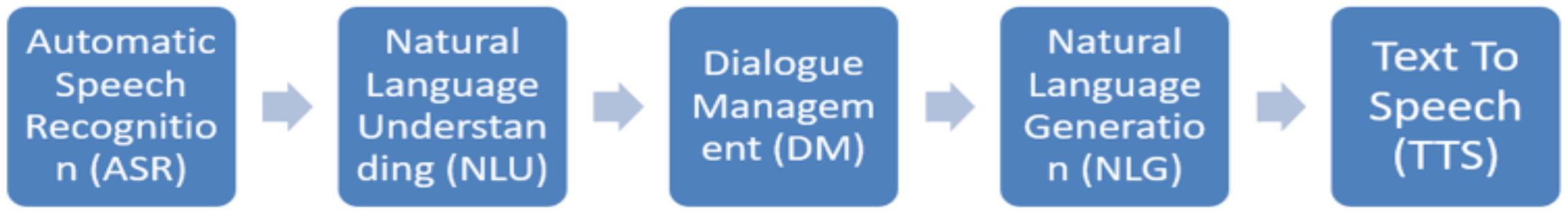

Figure 1. Flowchart of a typical chatbot system

Researches on rule-based matching chatbot were incited since the first chatbot was invented and tried the Turing Test in 1950s. To build a chatbot in such pattern require tremendous amount of human dialogues as knowledgebase. Moreover, this kind of simple chatbot for daily conversation was incapable to extract information to transform into knowledge and even generate new knowledge like agent with AI technology nowadays. Traditional chatbot with sufficient corpus can correspond to suitable responses for human questions in both grammar and matching rates due to responses are natural conversations produced by human. Additional matching words signified better selected responses. Thus, AI-based NLP technology challenges nowadays are machines' capability to generate responses rather than by patterns recognition which is the focus our language learning chatbot.

Neural network as language model in Natural Language Processing (NLP) supports machine to generate appropriate responses in recent years. Recurrent Neural Networks (RNN) with popular framework like TensorFlow and Keras are the mainstream for Language Model generation. In late 2018, Google published a basic language model called Bidirectional Encoder Representations from Transformers (BERT) with outstanding performance in 11 common NLP tasks which concentrated on Encoder scheme. Few months later, Open AI released another transformer based on unsupervised learning with pre-trained model focusing on Decoder scheme. By using unsupervised learning as pre-training scheme, the bi-directional transfer learning model can be served as a promising Language Model framework in NLP. With the pre-trained language model, our relatively small dataset can achieve better performance than traditional language models. Based on GPT-2 with fine-tuned model [1], our language agent has fluent and syntactic response as a virtual AI English tutor for industrial usage.

No matter of how excellent performance of these models, the essential neural networks are always in needed of big data as data source. Human minds make inferences that go far beyond the data available. The reverse-engineering of human learning and cognitive development helps the engineering of humanlike machine learning system [2]. Neural Network outputs are the mathematical computation results of neurons layers. It always considered as a black box, but the basic concept in bionics is inspired by human thinking and learning processes. We use the way of human learning and reasoning to explain the output of the neural networks, which also witness the development of search engines. That related to another question: How do human get information and knowledge?

Information system is the basis to build a knowledgebase, from websites to search engines, then to the ontology graph to retrieve the simplified output and make it more accurate. Due to the relation of keywords, the ranking done by search engines are more suitable for human justifications. In this paper, we use ontology, also called knowledge graph to simulate the connection of neural networks. The ontology graph is the tree of real-world concepts in different areas acquired by raw data, which focus on the relation between different nodes of the ontology graph. Just like neural networks, the tree also has the characteristic of synapses, which can inspire the relation extraction in ontology graph as memory in human brain. The interaction of agent with human also use natural language rather than query language of database or mathematical distance computation for similarity. Facing the barrier of machine can understand the natural language without computation, we use ontology graph to explain the humanlike neural networks. To some extent, the graph has ability to reason and generate new knowledge when it has sufficient knowledgeable and capable ontology graph that can "absorb" and "generate" new knowledge.

From the implementation perspective, English learning chatbot is constructed with Question-Answer-type of conversation as fundamental interactions between human and machine, in order to construct a humanlike English learning system. In general, such Question-Answer system with knowledge base is better than the system without database, such as Information Retrieval-based Question-Answer (IRQA) by crawler or search engine and chatbot with rule-based distance matching. The knowledge base is divided into two parts, the task-oriented knowledge aims at special-domain knowledge base like expert system. However, the chatbot for daily chats need open-domain knowledge to answer unpredicted questions. For example, the customer service chatbot like Ali Xiaomi [3], which is the typical example of E-commerce online support staff to substitute human online customer service. The more specific domain, the more suitable for chatbot to predict and set personality problems from users. The opposite is open-domain KBQA such as Siri for Apple, Xiao Ice for Microsoft, the interaction form provides a 24 hours personal assistant for users including database and APIs to search engine and other apps within one terminal to answer questions of open-domain knowledge.

In our English language learning chatbot system, we use unstructured data, English text, from daily dialogue to construct knowledge base with dictionary and graphs (ontology graphs) from fine-tuned dataset. With Python's AI ecosystem development platform, researchers will obtain more ideas between neural network and cognition to find a highly accurate answers from massive unstructured data.

From the implementation perspective, we propose a mini-program in WeChat for real-world usage, with the fine-tuned GPT-2 model [1] and speech recognition service from Google, whose three levels systematically English Learning method provide an efficient way in natural language learning. Simultaneously, the ontology graph visualized on Neo4j, graph database, to explain the generated response from agent.

The main contributions and originality of this paper include:

1. Following the Explainable Artificial Intelligence (XAI) concept to explain the output natural language from our Neural Network model.
2. The introduction of GPT-2 framework with dialogue format [1] as a language model for our system, different from the original GPT-2 used in text generation with reminders, to extend the usage of transfer learning into our language learning chatbot.
3. The successful integration of transfer learning substitute to traditional seq2seq model with ontology graph for the fine-tuning of dataset.
4. The creative idea in users' convenience to develop an AI NLP-based English agent into mini-program into WeChat as intelligent mobile English learning chatbot tutor.
5. The successful design and implementation of a Webchat-based mini-program for real-world use for English learning.

This paper is presented as follows: Section II is the literature review to review the contemporary chatbot system from technology companies such as Microsoft, Alibaba and Hugging face in respect of function design and technology component to analyse existing idea and optimize our idea for agent. Also, the research direction and related work for XAI and our practical method of using ontology graph with NLP. Section III states the framework and methodology in theoretical and practical of agent, which include THREE levels for English learning system for users, connectionism in language model with GPT-2 and ontology. Section IV presents the implementation of agent at mini-program and the testing of constructed system in real-world to analyse NLP raw data and interpreter, the ontology graph. Last section is to evaluate the performance combined with ontology for conclusion.

# 2. LITERATURE REVIEW

In recent decades, industrial market launched numerous chatbots, dialogue systems and online customer services. All of them are related products of Question-Answer (QA) systems. These systems have different technological backgrounds to control responses from systems. Furthermore, some researchers published structured dataset to fine-tune and evaluate the performance of systems to define the language model. Due to the particularity of our English language learning agent, we encountered situations that users require unstructured raw data as source text with the help of AI ecosystem to develop a multistep function-oriented learning method to language agent. This is a process from data pre-processing to the end of user interface to implement our idea for agent. Although AI technology provide lots of framework to generate conversation responses, we always adhere to research direction and related work of XAI and Ontology Graph (OG) with NLP.

## 2.1. An Overview of Chatbot

### 2.1.1. A Knowledge-Grounded Neural Conversation Model (seq2seq RNN) [4]

Since the origin of seq2seq model generated by RNN, neural network based chatbots had engaged in both industrial and academic communities. In 2018, Microsoft extended their industrial conversation system to make responses from the system to avoid brief and illogical contents as compared with human responses.

In respect of design, Microsoft already possessed functions to give simple responses in open domain. Figure 2 shows the extension exist only in the branch of Encoder to add facts into response. Both versatility and scalability in open-domain and external information knowledge of textual and structured are combined in this system, which has the recommendation system function for restaurant but not task-oriented.

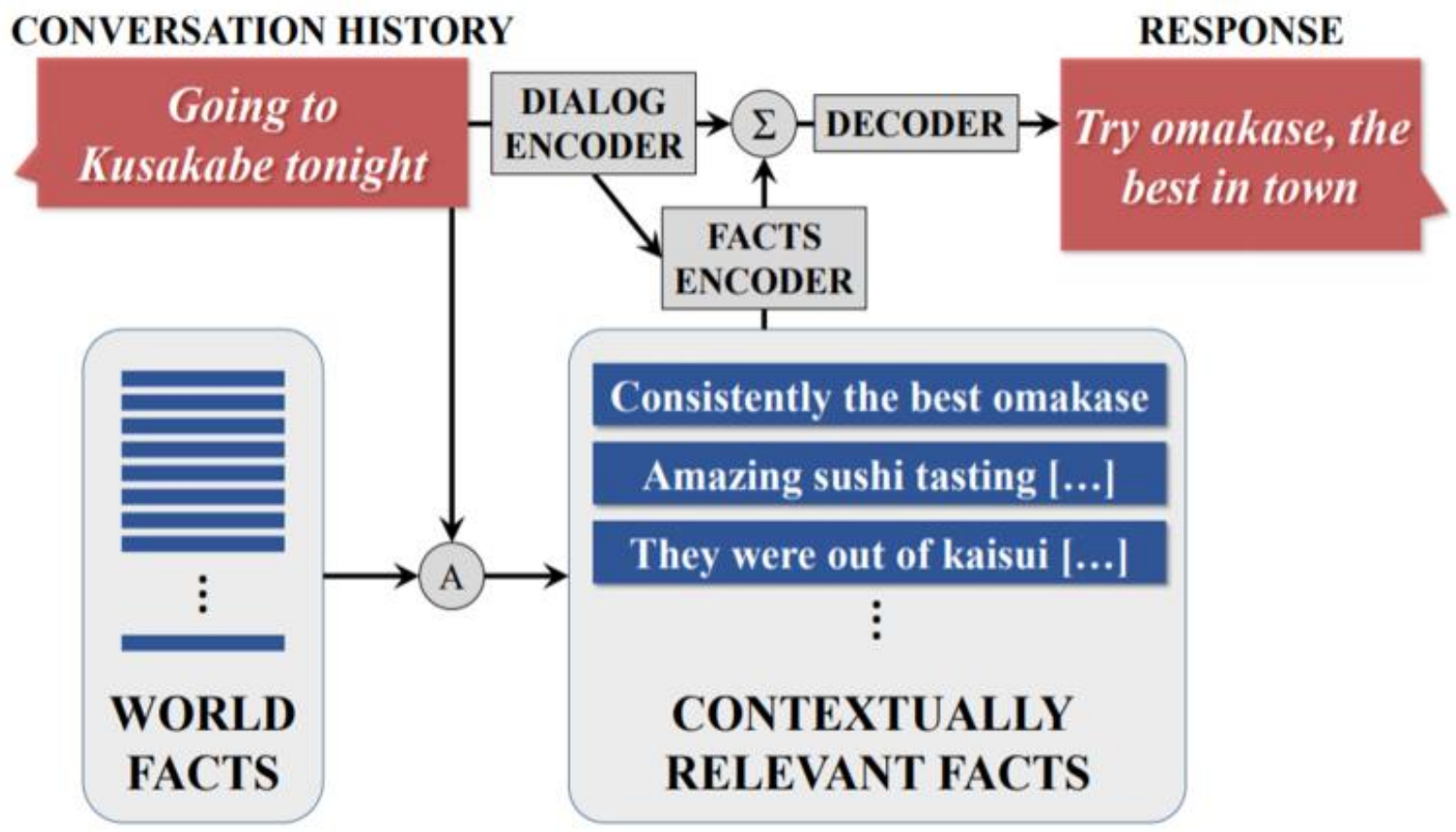

Figure 2. Architecture of Knowledge-Grounded Model

From the implementation aspects, different with the slot-filling to grounded content using rule-based scheme, the system uses another seq2seq neural network model fed by dataset from Twitter and Foursquare tips, the same technique with original language model. After that, the conversation provides with more meaningful and logical contents in responses, which only infuse knowledge information into the trained data-driven neural model.

The progress of knowledge grounded supported our NLP part to extract the Triple with useful information to consist knowledge. The meaning of XAI in our research is to explain the output response. This review shows the dataset can be absorbed by neural networks and generate output by sort or upgrade these data to information or knowledge, in which thinking resemble human to give response. In other words, abstract information can be visualized on Ontology and explained the response in relations.

#### 2.1.2. AliMe Chat: A Sequence to Sequence and Re-rank based Chatbot Engine (rule-based IR+seq2seq) [5]

In order to improve response quality to obtain the most matching response sentence within the restriction of neural network, AliMe, a commercial chatbot specialize in E-commerce industry, integrate both traditional information retrieval based and Seq2Seq neural models.

From the design perspective, AliMe is applied to E-commerce as a substitute of human service in Taobao. So, for the function of AliMe, it should search the most similar question to obtain response from the QA database to reply customers. If the client posed a new question to system, it should be intelligent enough to answer the question. At optimizer part, whether every response is generated or searched, both will be selected again by Seq2Seq neural network to obtain the most suitable response to human users. Also, AliMe is not limited in a task-oriented service for Taobao. According to survey by Alibaba, AliMe received 5% of questions within E-commerce span. It intended to upgrade to an open domain chatbot for questions expansion.

Before the transformation of AliMe to an open domain chatbot, most of responses depended on 9,164,834 QA pairs in database. After that, Seq2Seq model of GRU with SoftMax and Beam Search algorithms provided better response from neural network in both open and special domain. Also, the Seq2Seq model used second time to re-rank candidate responses.

#### 2.1.3. The Design and Implementation of Xiao Ice, an Empathetic Social Chatbot （IR+seq2seq+KG） [3]

Compared with chatbots mentioned before, Xiao Ice as shown in Figure 3 has the most complex chatbot structure built. The key-points of Xiao Ice is comprised of Intelligent Quotient (IQ) and Emotional Quotient (EQ) into the system design.

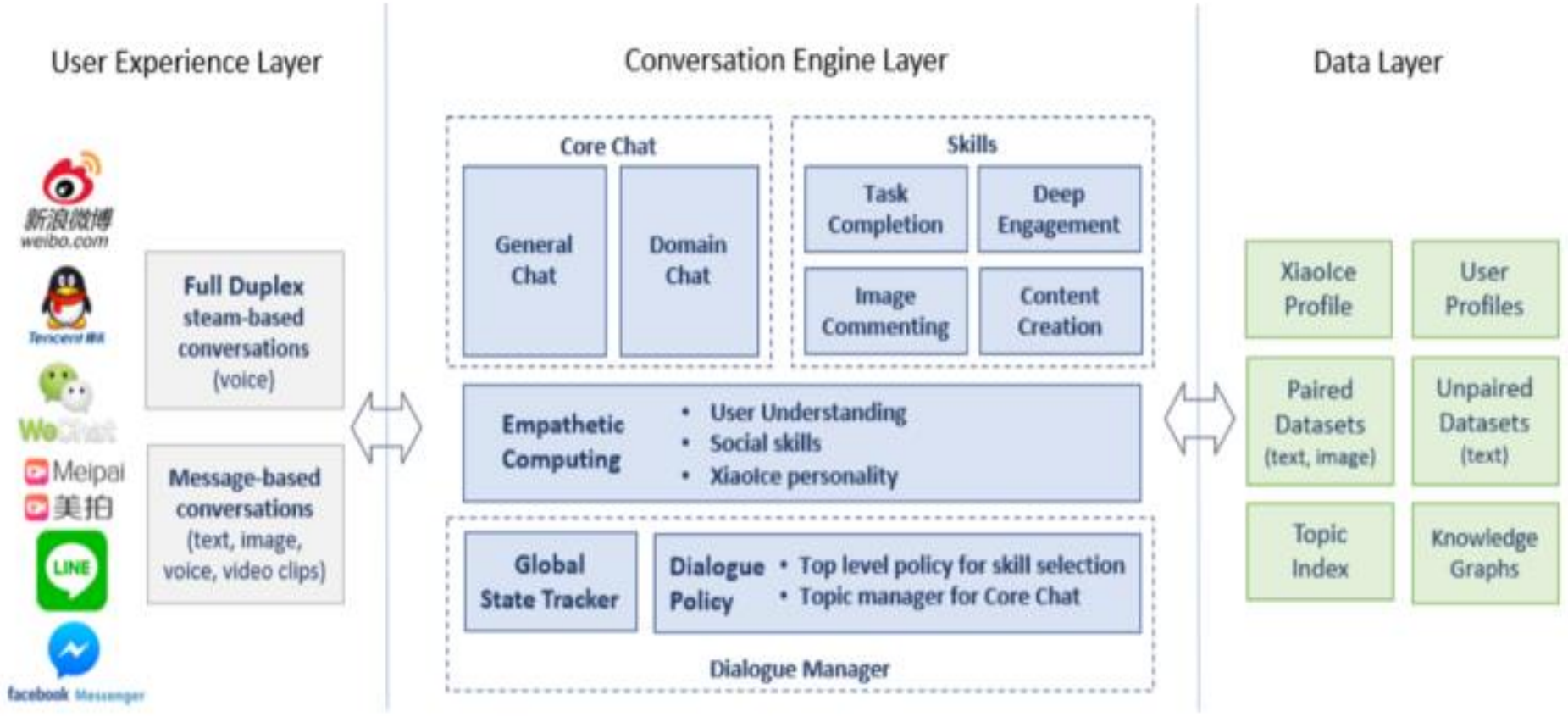

Figure 3. Xiao Ice System Architecture [3]

In the architecture above, Xiao Ice can adapt to input with different format, which not limited in voices and texts but processed images and videos. EQ and IQ show on topics adjustment with users' preferences in special and open domains. Considering at an English learning scenario, our agent defines the highest two layers for topic discussion and open domain of 'free-style conversation' to improve conversational skills aims to one field.

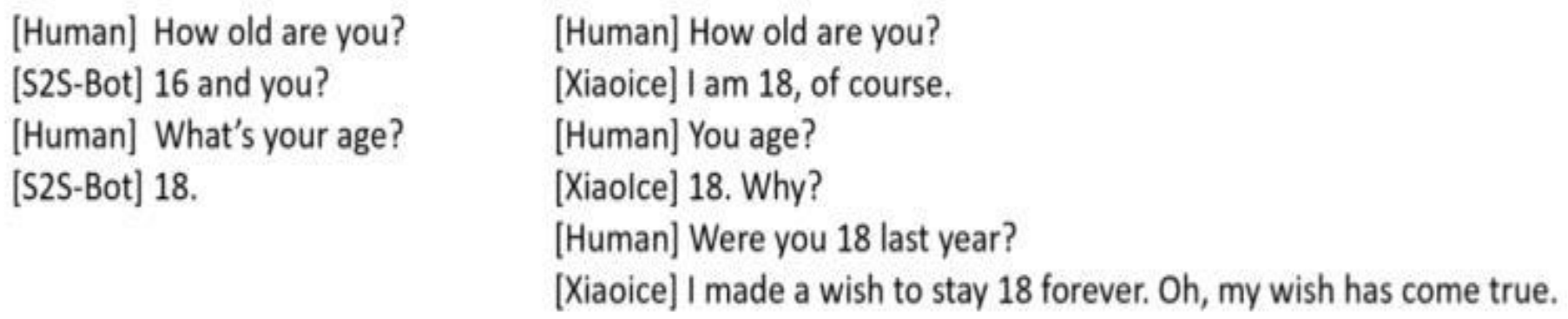

Figure 4. Examples of inconsistent responses generated using a seq2seq model [3]

The truth is, chatbot with emotion provided more natural and humanlike responses.

From the implementation perspective, the core of Xiao Ice is based on the language model using RNN to create responses. As mentioned, Xiao Ice integrates three methods to create responses. First, its chatbot has divided into retrieval-based and generated-based models. Xiao Ice has both candidate generator and candidate ranker. For the generator, it uses rule-based matching with real-world conversation collected and stored by natural language. Second, it uses deep learning model trained by paired dataset to simulate human dialogue to build human-like system. Third, it uses query on knowledge graph to get related entities. The candidate ranker also corresponding to the generator but including the semantic computation in NLP and empathy matching of Xiao Ice personality.

For our language learning agent, in order to simulate the IELTS test, we choose fine-tuned GPT-2 model with daily dialogue dataset to shape the tutor with different personalities and background. Our ontology graph acts as an interpreter for language model rather than graph database as compared with Xiao Ice.

### 2.1.4. Transfer Transfo: A Transfer Learning Approach for Neural Network Based Conversational Agents (transfer learning with GPT-2) [1]

The above chatbots discussed are basically adopting the Seq2Seq model, it gets good performance for the generation of responses. Since neural network is a data-driven model, it means that the performance are heavily relies on the amount and quality of the *big data*. Thus, based on attention mechanism, transfer learning using self-attention can used to find the relations and sequence dependency among words in the conversation.

The outstanding feature of transfer learning architecture in Google BERT and Open AI GPT-2 includes: 1) the proficient at encoder; and the proficient at decoder for response generation. GPT-2 uses self-attention method, which is an attention mechanism relating different positions of a single sequence to compute a representation of the sequence [6]. It extracts the relations of the sequence to rewrite the sequence. The process is similar to the extraction of information from text and make comprehension for further processing.

GPT-2 is more specialized in language generation according to self-attention scheme because it already absorbed 40G pure text to learn semantics and syntax of natural language. Fine-tuning is to use dataset with suitable format for special task to personalize the original GPT-2 model to task-oriented language model. Transfer Transfo we used as chatbot in our agent is a language system combining Transfer learning-based training scheme and a high-capacity Transformer model. [6] By using the persona-chat dataset to fine-tune the model, its utterance changes from long-text to dialogue format. Persona-chat in real-world helps to shape communicator's background to further define the topic and better understand user's input sentence. It shows the priority in the different AI language tutor with different personality resemble Speaking module in IELTS.

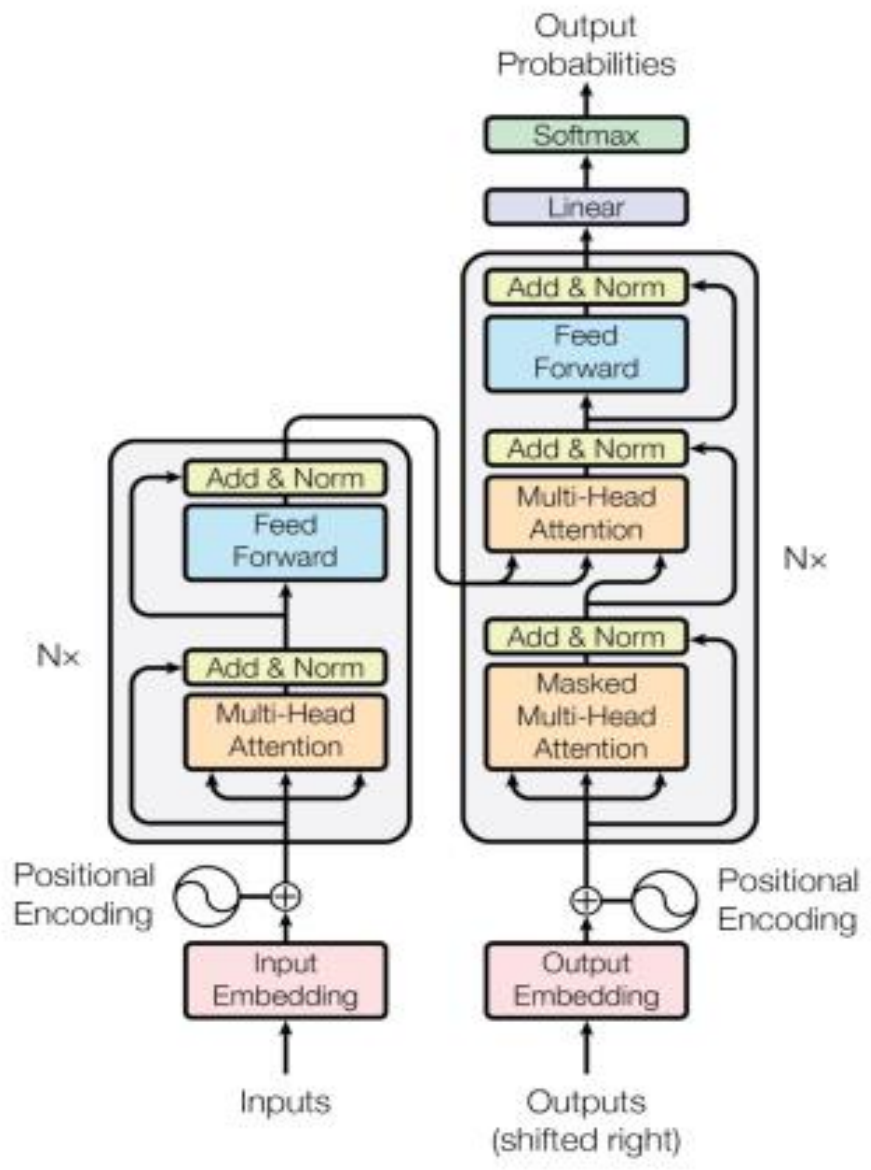

Figure 5. Transfer Learning-Model Architecture [6]

This paper shows the resulting fine-tuned model of significant improvements over the current state-of-the-art end-to-end conversational models like memory augmented seq2seq and information-retrieval models [6]. This is why we chose transfer learning rather than seq2seq model to apply XAI. The transfer learning achieved better and advanced performance, whereas self-attention is corresponded to our academic research in connectionism with ontology to explain the neural network are associated with bionics.

## 2.2. Explainable Artificial Intelligence (XAI)

Deep learning has widely applied in industrial projects. However, the explainable segment is the main barrier for artificial intelligence future development of. If the AI output model is unexplainable or uninterpreted, the output validity is unreliable. It is undeniable that AI is influential in prediction, classification and in decision making specifically the computation process of AI models is parameters, we only input big data for an output, but are unable to explain the recommended output response to human users. Just like the input format feed to neural model is word embedding of vector rather than natural language.

Neural network is a *black box* resembles to the construct of human brain. So far, we do not know its computation process. The process of human cognition is from concept to practice. It has axons and synapses to activate the storage of another neuron which is the human memory. The definition of classes, attributes (properties) are concepts and relationships as memory in human brain. The way to store these relationships is similar to human brain to transform raw data to knowledge, machine also requires special data format which is different with natural language to compute machinal thinking to reasoning new relations. So, we use triple <entity, relation, entity> as a basic format to store computer memory and visualize the connections by ontology. XAI is an interpreter to solve the question that the machine can understand the input data and get the output data human readable that means machine has ability to thinking like human. In NLP, it is Natural language Understanding (NLU). The NLU improvement is the progress of encoder part corresponding to end to end model. Natural Language Generation is the verification to get expected responds from human.

IBM delivered a speech on the Explainable AI (XAI) on April 2020 to visualize the implementation of the *true* Human-Computer Interaction (HCI). It meant the directional approach did not restrict to reveal the black-box algorithm but used user-centred approach to connect user needs and technical advancement.

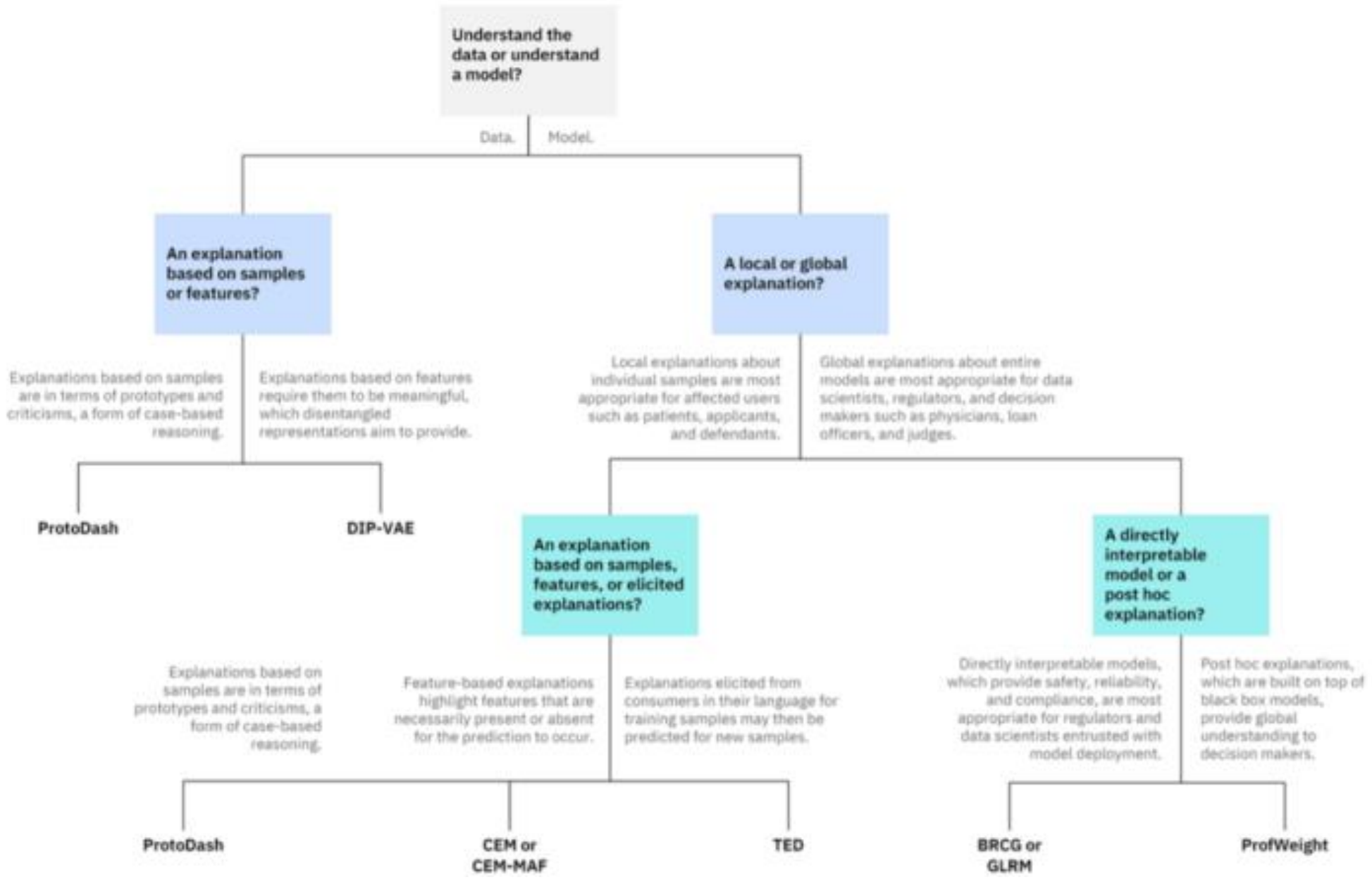

Figure 6. The taxonomy of AI Explainable methods introduced in AIX360 [7]

When we use AI technology on NLP, whether such algorithm can be trust and explainable is an important factor [7], which is also the core meaning of XAI. Thus, the algorithm operation should be able to understand and analysed by human. The following 5 covers the related XAI application. [8]

a) Transparency
   Focus on readable format by human
b) Causality
   Data-driven model provide both accurate inferences with decision background
c) Bias
   Black-box cannot calculate model bias against complicated computation
d) Fairness
   AI system always operate in a fair manner to users.
e) Safety
   AI system output are understandable and trustworthiness in regulatory sectors.

XAI transparency and compliance should be taken into account in association with a given related prediction [8]. The first approach is required more attention to deep learning and neural network considering it powerful prediction ability representing the deep explanation focusing on neural network layers and structures such as computation and parameters. The second approach is interpretable models such as casual models like linear regression, Bayes, logistic regression, decision tree are models in statistics which can be explained during calculation and reasoning. On the contrary, random forest, which consist of lots of decision trees has higher accuracy but are uninterpretable. Third approach is model induction which can infer any explainable model from black-box model. It put forward high-level requirements to explain every model for users against actual features. In this paper, we use the last method to interpret transfer learning-based dialogue system of end-to-end decoder part to obtain the response.

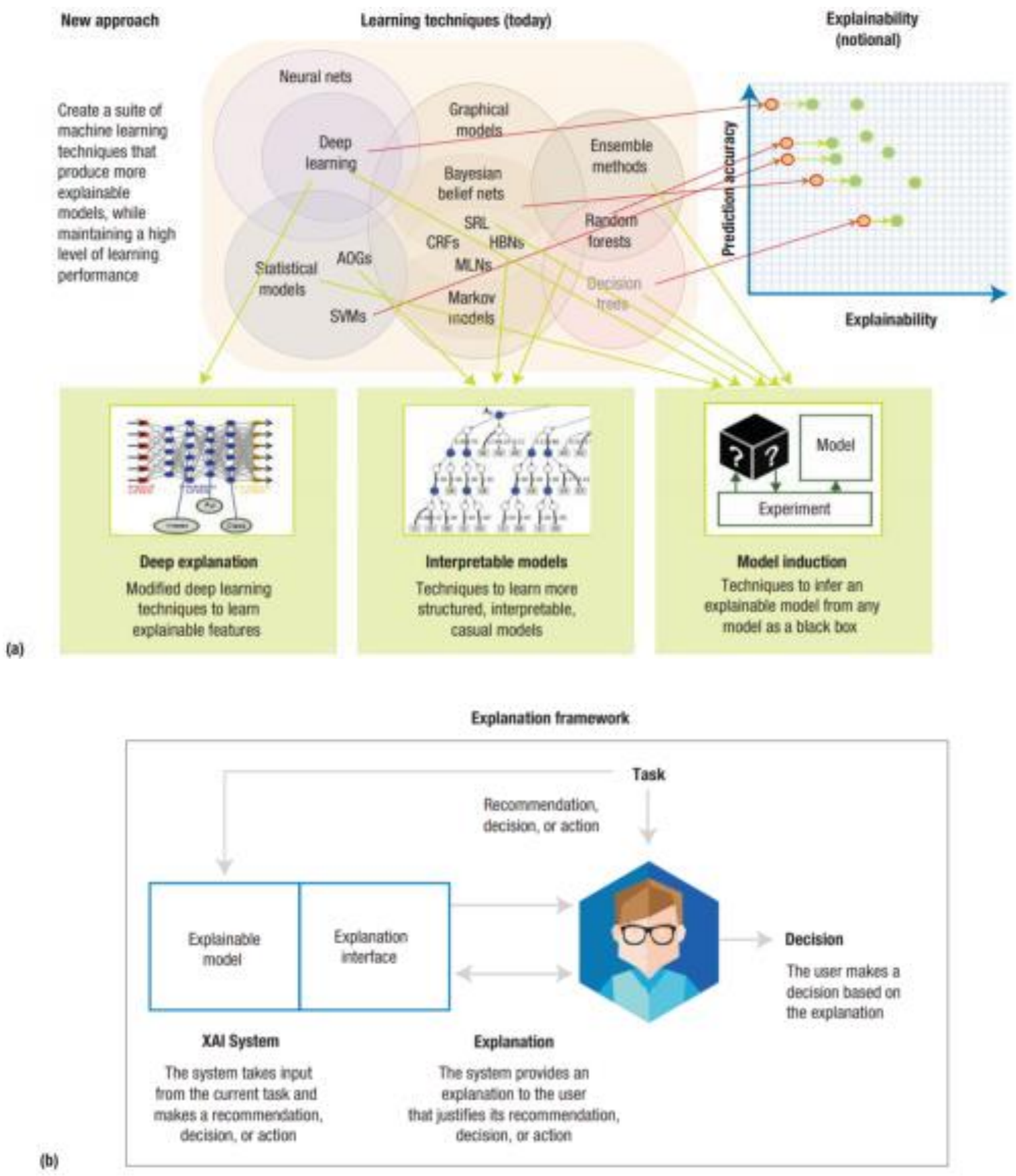

Figure 7. An overview of explainable models of AI (XAI).[8]

In this article [8], Hani Hagras explained the Fuzzy Logic Systems (FLSs) and human understandable AI, FLSs tried to mimic human thinking and research on the approximate way to thinking rather than limit human brain resemble to neural network. It has upgraded to the philosophy to build numerous of if-then reasoning rules to describe given human behaviour in human-readable way. The rules are the highest reasoning format for inference called OWL in ontology. Knowledge reasoning is a developing area in ontology and natural language generation. From existed knowledge to generate new knowledge based on logical rules to make up the contents of dialogue system is to be solved.

## 2.3. Ontology

Ontology could be seen as a visualization of knowledge base or the update of search engine, whose previous version is semantic Web. It represents the fundamental form of knowledge representation with graph. We are familiar with database, which used as the container to store different type of data. The knowledge is the senior of data. It should be extract from raw data such as domain texts or constructed database to save as the computer or human readable format, such as the triple of RDF, OWL. The hierarchy is shown in Figure 8.

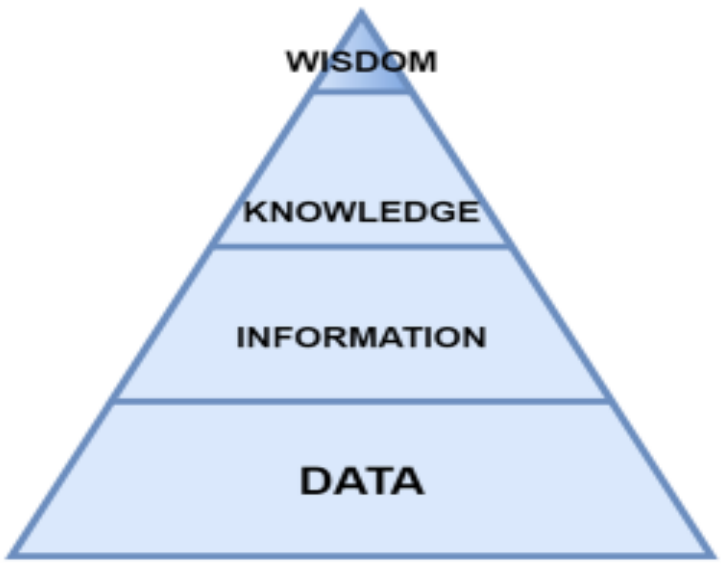

Figure 8. Pyramid of Knowledge

Ontology more like a tree to consist of the concepts, categories and relations systematically for the development of knowledge-based system, which responsible for question concerning about what entity exist and how such entity may be grouped and related within hierarchy and subdivided according to difference and similarities. [9] It is readable for both human and computer. Because the natural words are not transformed to the computer bits but use natural language processing tools to separate the objects sentence to triple format, which refreshed rapidly and interact between computers conveniently. Once we extract the triple as basic constructed data, then the graph database like neo4j, Orient DB, Amazon Neptune, will show the RDF/OWL triple to visualize the relationships of entities. The relations give ambiguous help to the inference of OWL to generate new knowledge.

Ontology has three levels, after the development of decade, another name from Google called knowledge graph, which also divide it into two level. For traditional ontology, the top-level corresponding to open-domain ontology graph (OG) and domain ontology match special-domain knowledge graph for specific industrial applications, for example finance, medicine. Domain ontology can extend to top-level ontology with grounded knowledge for concepts, because it built from bottom to top as open-domain to including as much as possible knowledge concepts for interoperable, information retrieval, automatic reasoning and other specific natural language processing tasks about common sense. In opposite, domain knowledge graph build from top to bottom, which always digs the deeper relationship to enlarge the domain ontology of different entities.

With the development of cognition and knowledge representation, knowledge engineering tools also update from handcraft to graph database. For academic, researchers hope the system really equipped with inference ability with OWL format, the tool proposed by Stanford University that protégé has updated to 5.5 but also half-manually to type the RDF/OWL triple. Protégé concentrate on special domain ontology edition owing to the handcraft, which is so hard to build open-domain knowledge. However, protégé supports various of ontology languages such as RDF, RDFs, OWL, XML, UML, etc. Figure 9 shows most ontology language and corresponding levels in semantic Web.

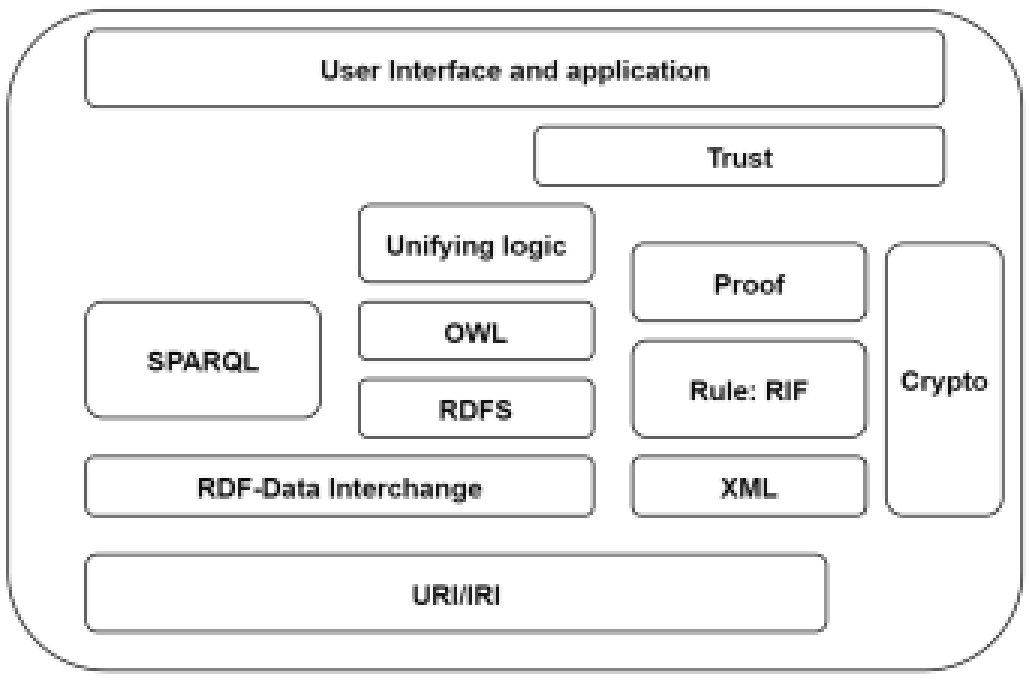

Figure 9. Semantic Web Framework

Now, knowledge engineering developers already built magnitude open-domain knowledge base for their purpose. Ontology designed to optimize the rank and contents of search engine also for commercial usage. It will re-rank the search output more intelligent through the keywords. Compared with traditional search engine, the relation can improve the search efficient and display related information.

The ontology storage system also develops with larger magnitude for commercial usage. For the vision and operation in construction, neo4j substitute other graph database or storage system to be the most welcomed graph database to visualize and analysis by Cypher, a custom-made SPARQL for neo4j. Figure 10 is the rank of storage system in September 2020.

☐ include secondary database models    32 systems in ranking, September 2020

| Rank | | | DBMS | Database Model | Score | | |
|---|---|---|---|---|---|---|---|
| Sep 2020 | Aug 2020 | Sep 2019 | | | Sep 2020 | Aug 2020 | Sep 2019 |
| 1. | 1. | 1. | Neo4j | Graph | 50.63 | +0.44 | +2.41 |
| 2. | 2. | 2. | Microsoft Azure Cosmos DB | Multi-model | 31.67 | +0.94 | +0.80 |
| 3. | 3. | ↑ 4. | ArangoDB | Multi-model | 5.80 | +0.06 | +1.44 |
| 4. | 4. | ↓ 3. | OrientDB | Multi-model | 5.48 | +0.47 | +0.42 |
| 5. | 5. | 5. | Virtuoso | Multi-model | 2.56 | -0.09 | -0.24 |
| 6. | 6. | ↑ 7. | Amazon Neptune | Multi-model | 2.35 | +0.20 | +1.15 |
| 7. | 7. | ↓ 6. | JanusGraph | Graph | 2.34 | +0.32 | +0.89 |
| 8. | 8. | ↑ 17. | FaunaDB | Multi-model | 1.86 | +0.15 | +1.43 |
| 9. | 9. | ↑ 10. | Dgraph | Graph | 1.61 | +0.15 | +0.65 |
| 10. | ↑ 11. | ↑ 13. | Stardog | Multi-model | 1.45 | +0.09 | +0.73 |

Figure 10. DB-Engines Ranking of Graph DBMS, top 10, date Sep 2020 [10]

The rank of DB-Engines graph system shows the popularity degree of different type storage system. Considering the running speed to generate graph relations and storage format extracted from raw data, neo4j always the top-1 causing it industrial magnitude and relative high refresh speed, more important is the API with python, java and other frequently used language.

## 3. METHODOLOGY

To start with, we need to define some questions for our English Language Learning agent first. We prefer to define it as an QA-based dialogue system rather than for task-oriented execution but with grounded knowledge and special scenario. Combing with XAI, our system should concern three aspects and make direction for theoretical research. [11]

The first aspect is artificial intelligence (AI). Once AI mentioned, from the aspect of understanding of human things, it has two layers, the content and the methodology. In general, we divide AI into two mainstreams, Computer Vision (CV) and natural language processing (NLP). The other hand, methodologies to research AI is Symbolism, connectionism, and behaviourism.

Symbolism is an intelligence that use mathematical logical thinking to simulate the human thinking. For example, the expert system. Connectionism belongs to bionics for human brain, in general, neural network is a typical model. Behaviourism focus on the prediction of human behaviours, such as AlphaGo of Reinforcement learning and genetic algorithm, in which researchers think human get adaptivity from the interaction with external environments. In our research, rule-based system is symbolism but generate-based with word2vec is connectionism by neural network.

The second aspect is natural language, which is truly the real-world human communication language. NLP is relative to computer. It hopes computer can understand human language.

The third aspect is understanding. The branch of NLP in understanding is Natural Language Understanding. For human, there are a thousand Hamlet in a thousand people's eyes, which means the understanding related to similar life experience, common topics, context, knowledge base of individual and semantic and pragmatic of sentence more than dictionary. So, our direction to integrate as more as possible about what human thinking factors to applied in human intelligence for machine.

For XAI, readable and interpretable for human is explainable AI techniques. If the machine concerns these factors like human, no matter the computation method, the output and parameters can be interpreted by human trustily. In practical, our system research following the connectionism. By the reviews about self-attention mechanism, the ontology with connection are similar to the relations of attention mechanism.

In order to develop a complete system, we first define the direction as an AI English tutor with three levels. That part is pre-set at the User Interface (UI) based on mini-program in WeChat developer. It means WeChat user can log in and use it directly. As a platform, mini-program need whole NLP architecture to process the learning tasks. Our architecture has five parts. The first and last belongs Voice Recognition, which part we use service from Google, Audio to Words. NLU and NLG use the model of fine-tuned use daily dialogue from thesis Transfer Transfo: A Transfer Learning Approach for Neural Network Based Conversational Agents. [1] based on Open AI GPT-2.

As shown in Figure 11, the central part, Dialogue Management (DM), we use ontology graph for visualization on neo4j platform, where Ontology will explain the relations of entities in fine-tune data and generated response from language model. We will display the techniques we used in our system below, that Spacy for NLP raw data processing, neo4j as graph database to store ontology. Open AI GPT-2 as the original model to specific used into dialogue generation.

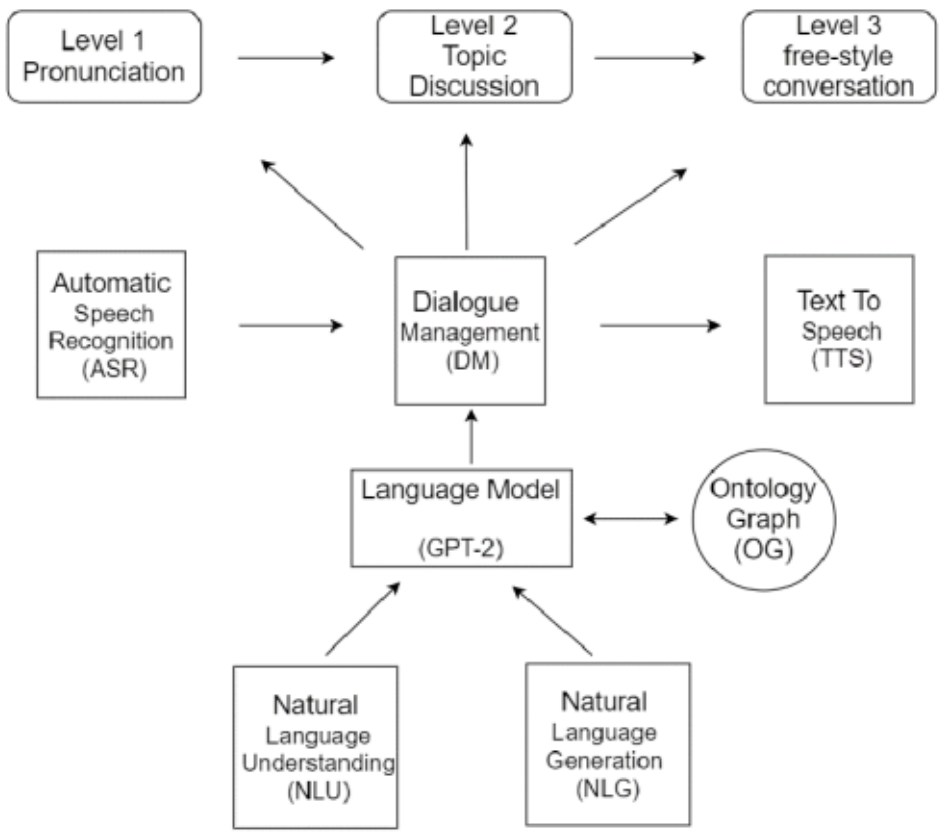

Figure 11. Architecture of this paper

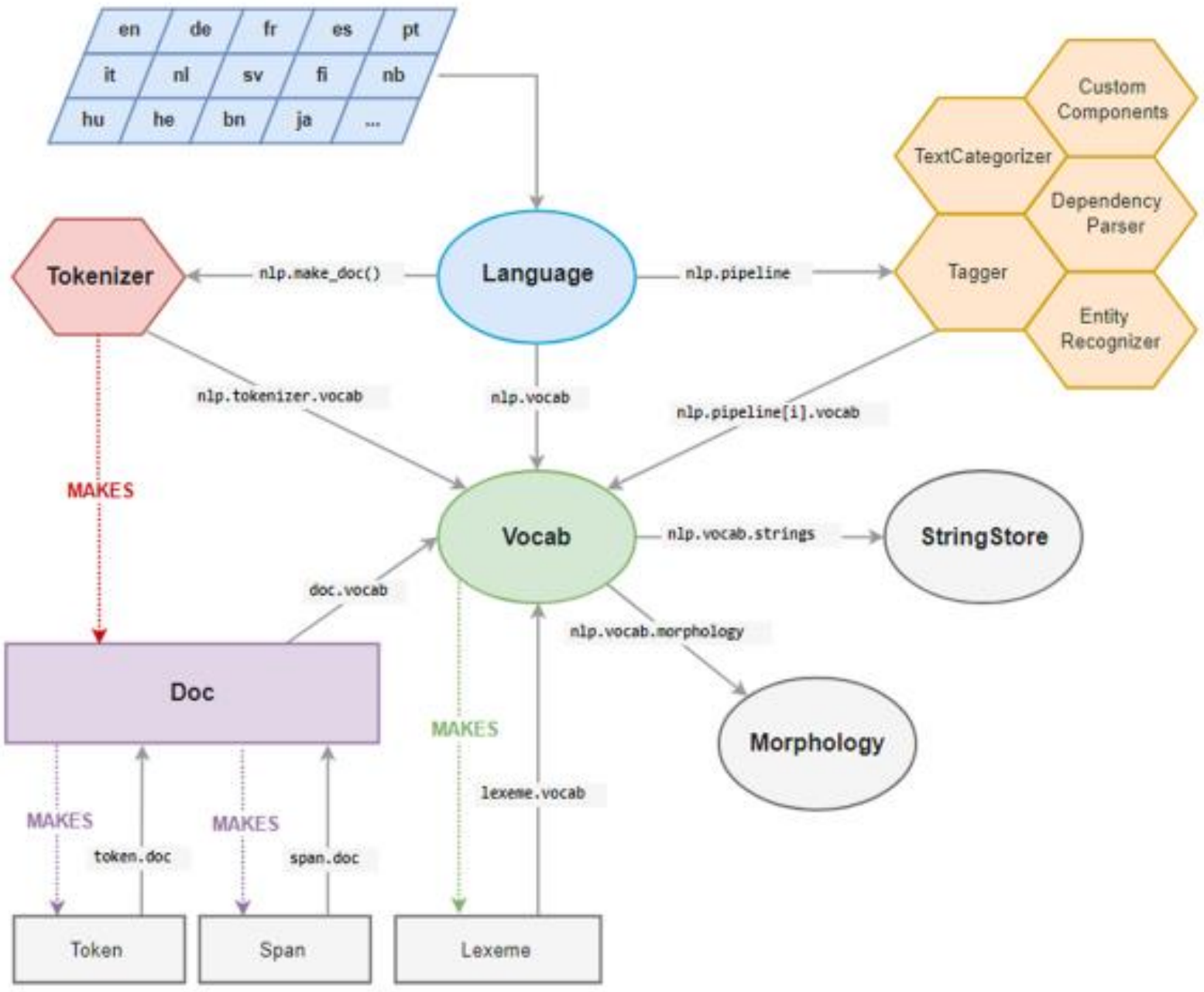

Figure 12. Architecture of Spacy [13]

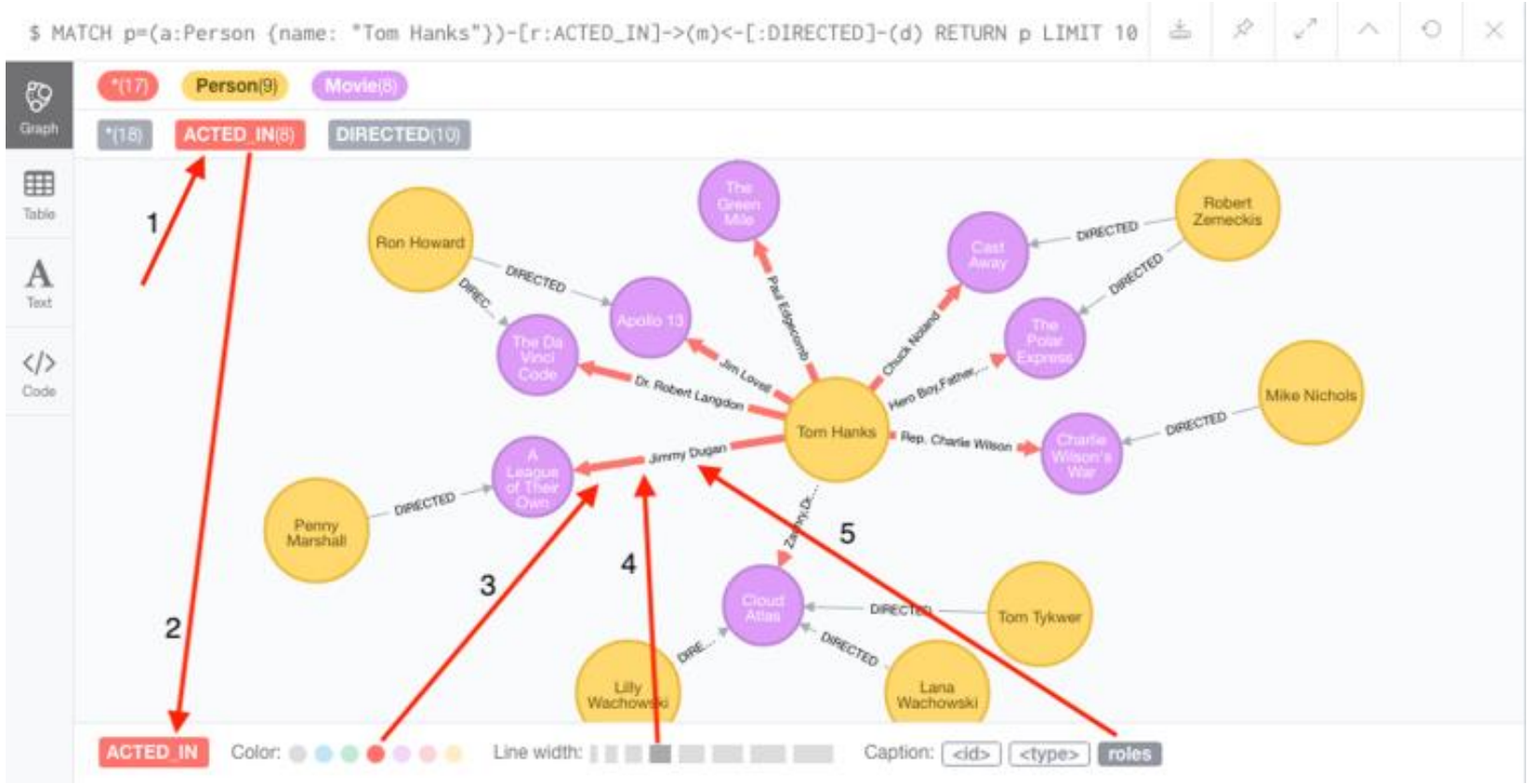

Figure 13. Demo of Ontology in Neo4j [14]

# 4. SYSTEM IMPLEMENTATION

## 4.1. Chatbot at Mini-program in WeChat

Before we explain the output sentence, following XAI, we integrate the language model, the GPT-2 model-based Transfer Transfo [1] with our THREE-level functions for language learning. Level 1 is pronunciation correction, where match the difference between standard pronunciation and human interaction. Level 2 is Topic discussion by pre-define the topics into UI.

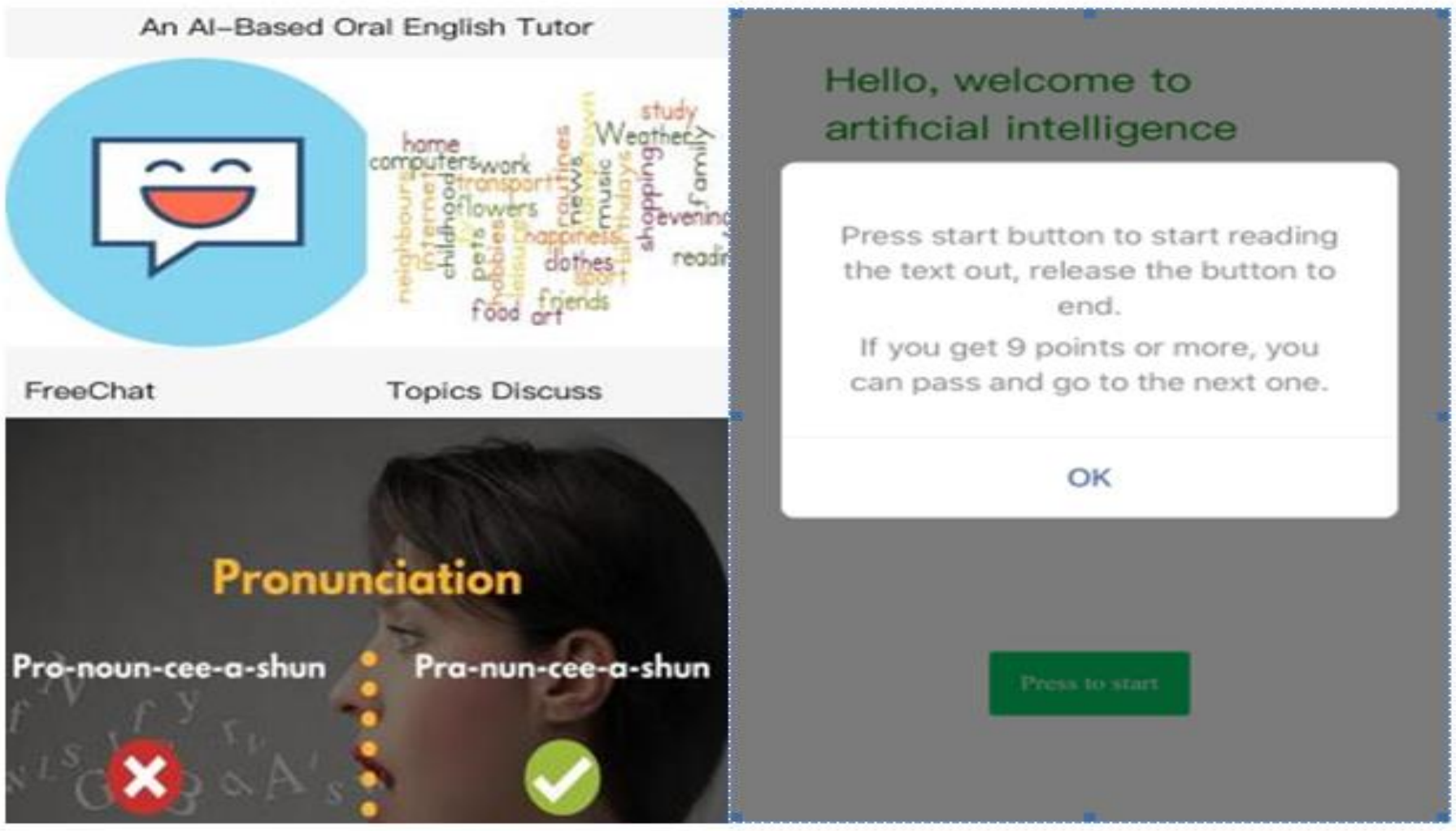

Figure 14. UI of AI English tutor develop by ourselves.

Level 3 is free-style conversation, which also the key-point part to be explained by Ontology. The Free-style conversation module has no restriction, which use the fine-tuned GPT-2 model to generate response. The system, from the perspective of industrial usage like this:

Step1: Record the user's voice and store it in a local temporary file.
Step2: Upload temporary file to server to analysis and convert it to text.
Step3: Use the text as input string to our language model to get the generated response back to front-end (mini-program).
Step4: Show the response text in the Chat Room.

## 4.2. Ontology Graph

Different with protégé which only could be create the entities and relations by hand. Neo4j take in so many kinds of data format. So, we cooperate neo4j with python to create the nodes and links automatically. The dataset of Open AI GPT-2 is json format, so we first transform the json data to txt data as text for entities and relations extraction. Then we use Spacy web model to extract the SPO triple as our knowledge. Every entity contains a property or an extended range to expand knowledge to get more information based on web. So, we choose DBpedia as our entity links. Because it is the constructed data from Wikipedia, which already an OWL database. It will help our ontology graph link to the internet semantic web to enlarge our database and extend the entities. After we have extracted SPO triple, we should build the node and relations for the neo4j according to it.

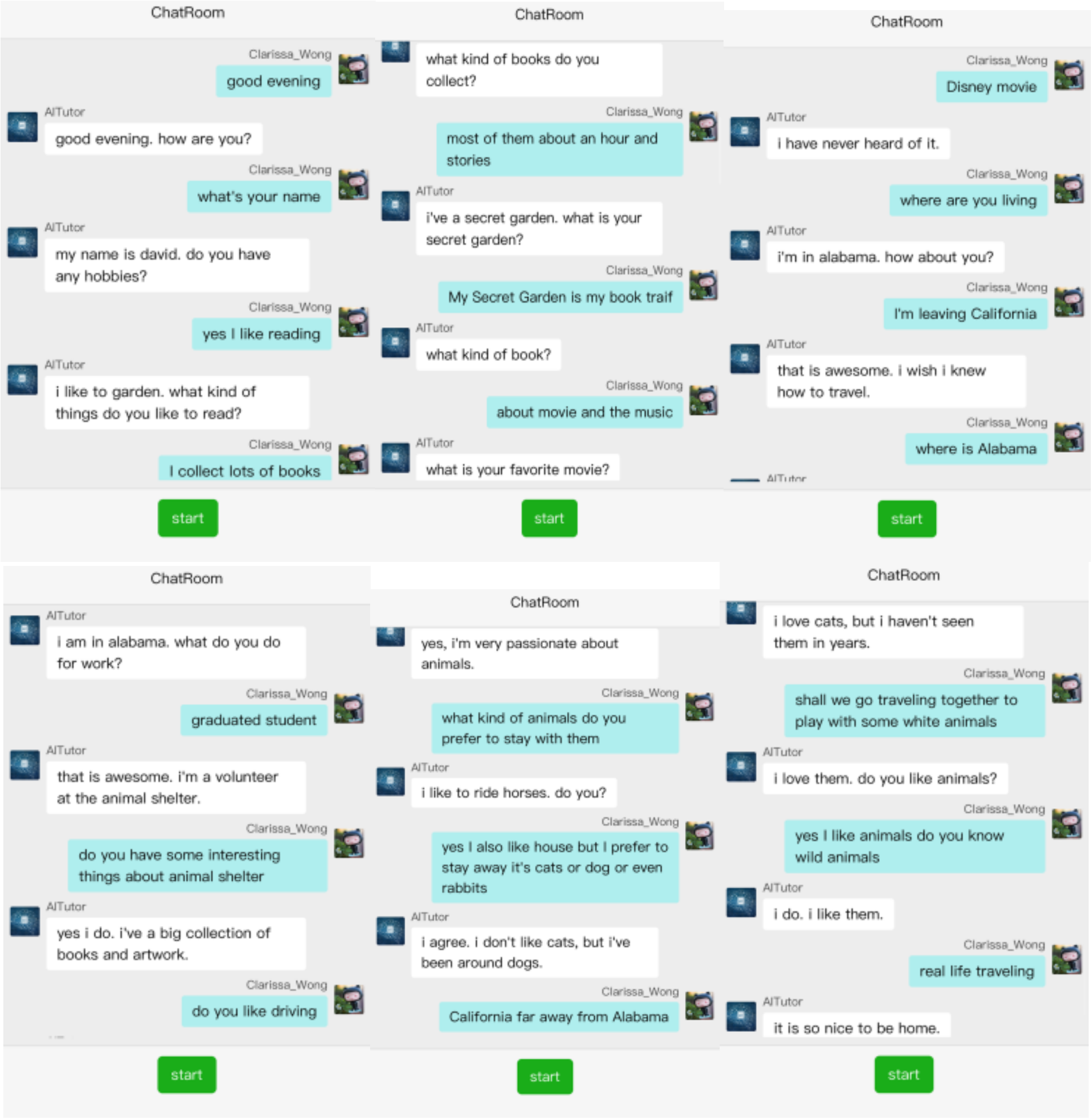

Figure 15. Test of Level 3 free-style conversation.

The conversation is the Level 3 free-style conversation, machine with back-end GPT-2 can generate any possible response as feedback. With this system, first three sentences are general speaking about hobbies in music and reading. When we talk about work, the contents from general to special domain about animals. We find that the special noun ‘animal shelter’ in part of training data has relatively high IDF (inverse-document frequency), so the TF-IDF score is higher. We find the most similar paragraph with dialogue content of the response we show in Figure 16.

I am from New York. where are you from? Well, good luck and hope your dream comes true!! I am good. how are you. I am sorry , I didn't get your name . I am Mary jerry, do you like animals? I work with them. hey. having a good day? and I am getting a dog very soon mostly in state things, i don't really get chances to go places. I am in school, and I volunteer at an animal shelter. hello, I am enjoying some crisp country air. what about you? I love snakes, I just read a book about snakes recently! yeah, I am quite busy too. hello, I am an attorney. hi there. how is it going? that makes complete sense. gotta go where the jobs are. I can't do fast food. my grandmother lives in my pool house. oh, okay. do you have any recommendations on shows to watch on Netflix? that is so nice! I wish you luck. personality. I go to at least 10 concerts a year. I work in retail. Madonna is my all-time favorite. lady gaga is my current favorite singer. hey, how are you? just got back from a long walk, so I am beat. Well, me and the wife and kids love traveling in my spare time. wow, that s awesome! in feel with you. I drive an old dodge it still runs pretty well. do you have any sisters? Oh, wow I bet you have to talk to people all the time that would be hardi enjoy taking care of my horse?

Figure 16. Sample text to Ontology.

We choose the matching sample text from the fine-tune dataset of GPT-2 model and transform the format of .json to the natural language, which will be the input to be extracted triple with Subject-Prrdicate-Object form.

```
[['I', 'am', 'attorney'], ['that', 'makes', 'sense'], ['you', 'have', 'recommendations'], ['I', 'wish', 'you'], ['Madonna', 'is', 'favor
ite'], ['lady gaga', 'is', 'singer'], ['I', 'drive', 'dodge'], ['you', 'have', 'sisters'], ['you', 'have', 'to talk'], ['i', 'enjoy', 't
aking']]
```

Figure 17. Extracted Triple.

In order extract the text, textacy is another package which is built on Spacy for SPO triple. After the Spacy deal with its pipeline tasks of the model to pre-processing the data for triple extraction, such as tokenizer, Tagger, lemmatization, coreference resolution, textacy will extract the SPO tiple and print all of them as list. The result display at figure17, which is the output of text figure 16 [12]. Compared the original text and SPO triple, even though the text following the S-P-O syntax, there are also amount of entities cannot be distinguished by Name Entity Recognition in Spacy. With my experience of protégé of Stanford University to build ontology graph by hand, we decide to add the absent triple which sentence included in the demo paragraph with the Cypher commands in neo4j. The syntax of English has five basic sentence patterns: S-P-O, S-P-O-O, S-P-O-Object complement, S-intransitive verb, subject-linking verb-predicative (SVP). The handcraft triple will obey these five basic rules. Go through all triples corresponding to original sentence, we add the DBpedia link to the entities, which existing in the DBpedia with 'url' link to extend knowledge base.

After the automatic recognition of SPO triple, with 17 relationship types link to different two entities, the most subjects are 'I', due to the training dataset is daily dialogue with special personality in US. Even though the 40G pre-trained dataset of Open AI GPT-2, it only equips the transfer learning model with fundamental grammar to generate suitable response with correct syntax. However, the contents and the length of sentence are controlled by fine-tuning part of Transfer Transfo with daily dialogue dataset.

I am from New York. where are you from? Well, good luck and hope your dream comes true!! I am good. how are you. I am sorry , I didn't get your name . I am Mary jerry, do you like animals? I work with them. hey. having a good day? and I am getting a dog very soon mostly in state things, i don't really get chances to go places. I am in school, and I volunteer at an animal shelter. hello, I am enjoying some crisp country air. what about you? I love snakes, I just read a book about snakes recently! yeah, I am quite busy too. hello, I am an attorney. hi there. how is it going? that makes complete sense. gotta go where the jobs are. I can't do fast food. my grandmother lives in my pool house. oh, okay. do you have any recommendations on shows to watch on Netflix? that is so nice! I wish you luck. personality. I go to at least 10 concerts a year. I work in retail. Madonna is my all-time favorite. lady gaga is my current favorite singer. hey, how are you? just got back from a long walk, so I am beat. Well, me and the wife and kids love traveling in my spare time. wow, that s awesome! in feel with you. I drive an old dodge it still runs pretty well. do you have any sisters? Oh, wow I bet you have to talk to people all the time that would be hard. I enjoy taking care of my horse? ↵

Figure 18. the SPO sentence unextracted by Spacy.

The coloured text is the part of text, which match with the SPO but not recognized by Spacy. With my experience of protégé of Stanford University to build ontology graph by hand, we decide to add the absent triple which sentence included in the demo paragraph with the Cypher commands in neo4j. After we extract the triple from text manually, the neo4j will display more 12 relationships with almost same subject entities. Because the dialogue always uses personal pronoun.

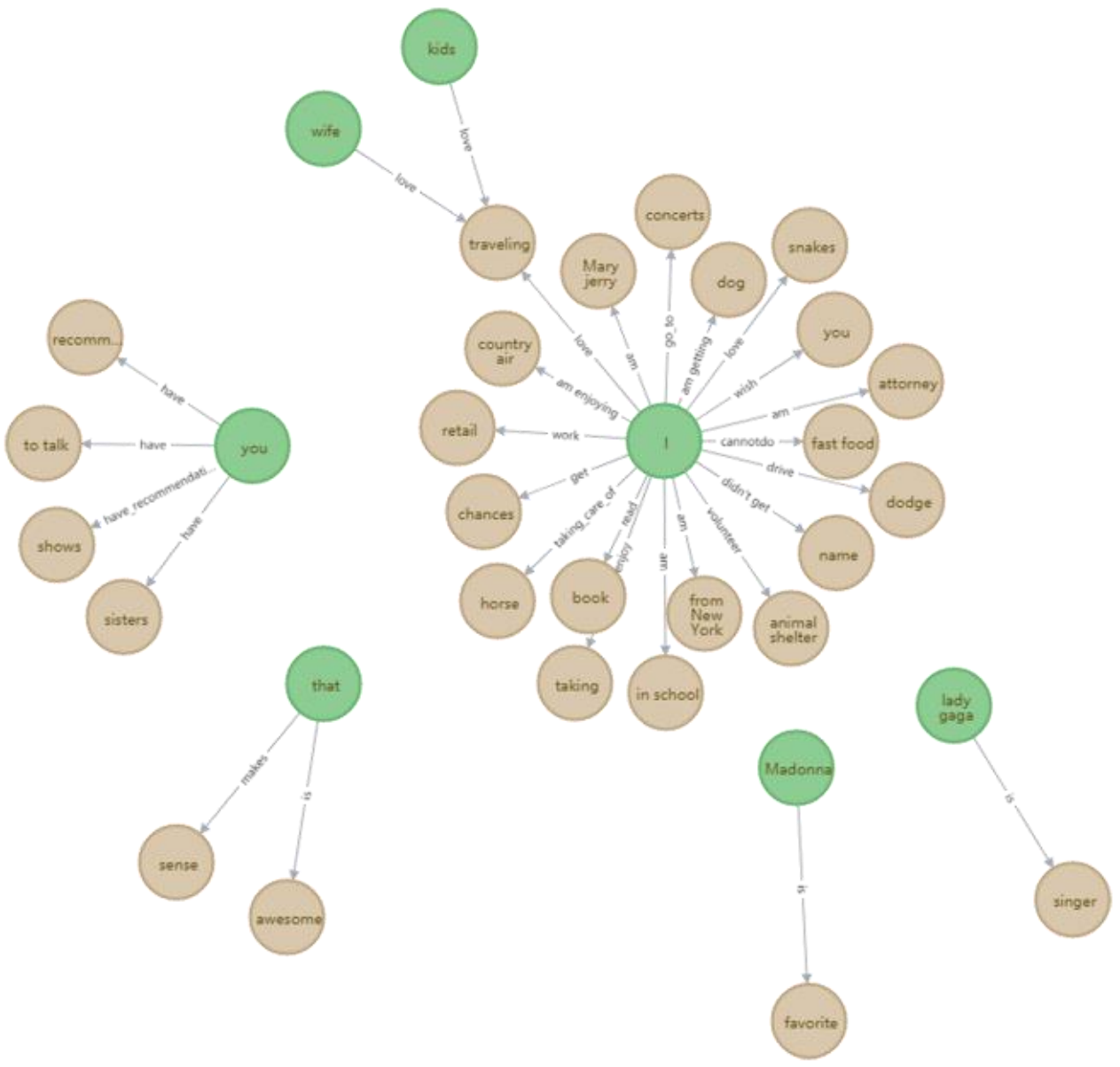

Figure 19. ontology graph with whole triples

In training dataset sample, it shows the paragraph we extracted 30 relationships also contains the content of dialogue. Such as the animals of horse, volunteer at animal shelter.

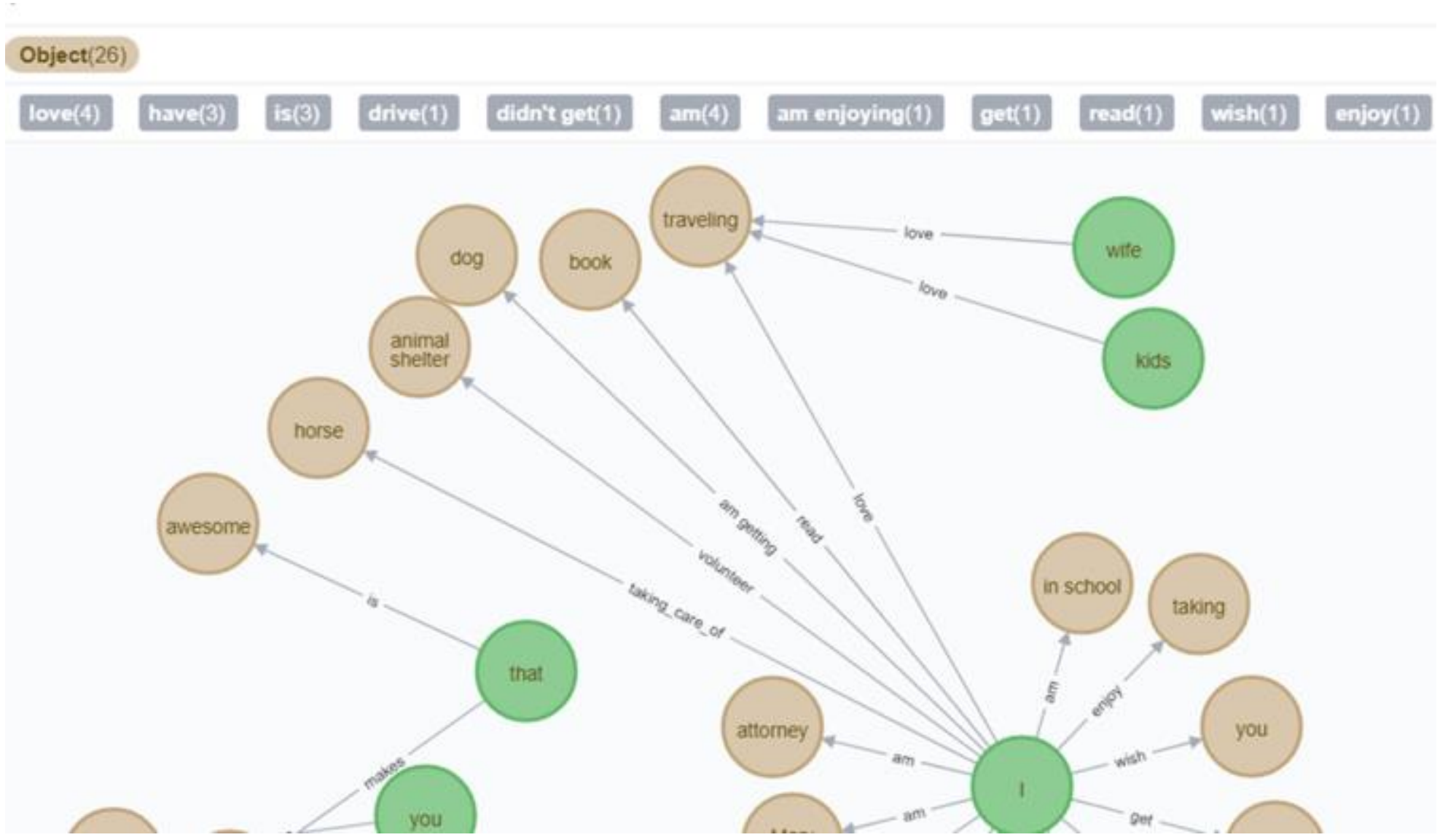

Figure 20. ontology graph visualizes the generated response

In order to prove the output sentence from system is explainable, we use some extract related triple in this demo paragraph. Due to our dialogue based on whole training data, the 200M json format data for trained. In general, only the rule-based pattern recognition can match the high-similarity sentence in database as response. However, the transfer learning and Transfer Transfo model are trained on end-to-end based neural network, which consider as a black-box.

Table 1. comparison of system generated and SPO triple

| system generated | training data triple at ontology graph |
|---|---|
| I'm a volunteer of the animal shelter | I volunteer at animal shelter |
| I like to ride horses | I taking care of horse. |
| I 've been around dogs | I am getting dog |
| I have a big collection of books and artwork.<br>( the reminder from user has collect) | I read book |
| that is awesome | that is awesome |
| … | … |

The specialization is that the Subject entities (the brown bubble) is too little compared with Objects (green bubble). From the aspect of our research direction connectionism, neural network is the bionic research, our ontology belongs to philosophy but the construct resembles the connection in humans' brain. As we see, the green bubbles represent different entities, but due to the same object, it will activate by the objects. At the beginning of this paper, we mentioned that we use ontology graph simulate the neural network. The partly visualization of training data shows the partly explain the output sentence and simulate the relationship in human's brain.

The size of ontology graph and language model restricted by the hardware. If the ontology extended with the whole training data, to some extent, with the help of TF-IDF for special words, we can explain more about the black-box.

## 5. CONCLUSIONS

In this paper, we design and implement an English Learning chatbot with the theory of explainable Artificial Intelligence using Ontology Graph (OG) and Transfer learning. We apply connectionism both in neural network and ontology into the simulation of human brain. In practical, our system techniques refer to Natural Language Processing (NLP) especially in Natural Language Understanding (NLU) and Natural Language Generation (NLG), Ontology, Transfer Learning, Search Engine and WeChat developer of mini-program. Due to the project is an industrial project, at experiment, we test several times with the three level to make sure that the system can be used to English Spoken training in a silent environment. From research aspect, our idea that use Ontology Graph to explain the output natural language sentences make a little progress. It means the neural network model except the feature that Open AI GPT-2 generate text according to the reminder written in algorithm, besides, the content of output sentence can be explain and visualize at the ontology graph with the same training dataset. The larger ontology graph contains more training data, the better and detailed explainable of output sentence.

### ACKNOWLEDGEMENTS

The authors would like to thank for UIC DST for the provision of computer equipment and facilities. This project is supported by UIC research grant R202008.

## AUTHORS

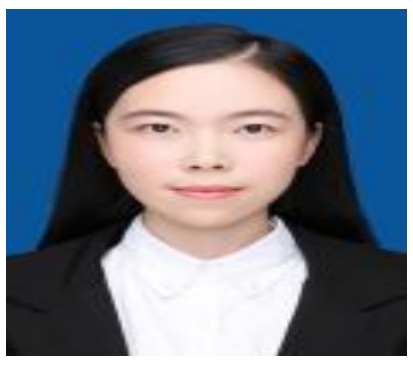

**Clarissa N. B. Shi** is graduated from Beijing Normal University - Hong Kong Baptist University United International College (UIC) awarded M.Sc. (Data Science) with distinction from Hong Kong Baptist University in 2020. Her research interests covering Ontology graph and Chatbot in Natural Language Processing, Neural Network with Mathematical and Explainable Artificial Intelligence (XAI).

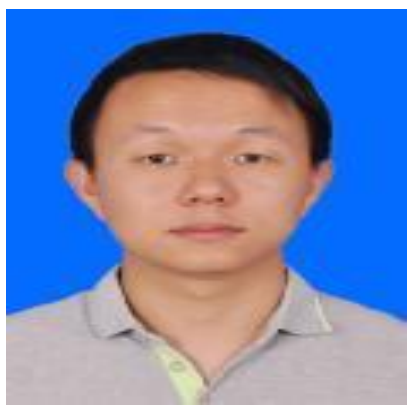

**Qin Zeng** is graduate from Beijing Normal University-Hong Kong Baptist University United International College (UIC) awarded his M.Sc. (Data Science) from Hong Kong Baptist University in 2020. Qin Zeng had worked in the software industry for decade before obtaining the master degree and has a very rich experience in Linux environment, cross-platform, multi-language development and deployment. His research interests covering natural language processing, recurrent neural networks and intelligent mobile applications in WeChat platform.

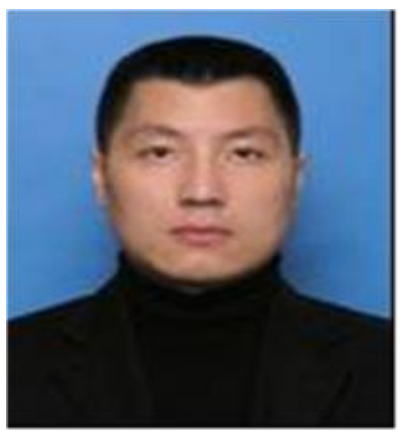

**Dr. Raymond S. T. Lee** (M'98) attained his B.Sc. (Physics) from Hong Kong University in 1989, M.Sc. (IT) and PhD (Computer Science) from Hong Kong Polytechnic University in 1997 and 2000 respectively. Dr Lee had worked at the Department of Computing of Hong Kong Polytechnic University as Associate Professor 1998 - 2005. During the past 20 years, Dr. Lee has published over 100 publications and the author of 8 textbooks and research monographs covering the fields of artificial intelligence, quantum finance, e-commerce, pattern recognition, intelligent agents and chaotic neural networks. Dr. Lee is now the Associate Professor in Beijing Normal University-Hong Kong Baptist University United International College (UIC) working in the field of quantum finance, quantum anharmonic oscillators, chaotic neural oscillators, fuzzy-neuro financial systems, chaotic neural networks and severe weather modelling and prediction.